\pdfoutput=1

\documentclass[11pt]{article}

\usepackage{EMNLP2022}

\usepackage{times}
\usepackage{lscape}
\usepackage{algorithm}
\usepackage{algpseudocode}
\renewcommand{\algorithmicrequire}{\textbf{Input:}}
\renewcommand{\algorithmicensure}{\textbf{Output:}}
\usepackage{amsmath}
\usepackage{amsfonts}
\usepackage{amssymb}
\usepackage{tablefootnote}
\usepackage{latexsym}
\usepackage{booktabs}
\PassOptionsToPackage{table, xcdraw}{xcolor}
\usepackage{float}
\usepackage{graphicx}
\usepackage{enumitem}
\usepackage{amssymb}
\usepackage{array}
\usepackage{tabularx}
\usepackage{caption} 
\usepackage{physics}
\usepackage{tikz}
\usepackage{mathdots}
\usepackage{yhmath}
\usepackage{cancel}
\usepackage{color}
\usepackage{siunitx}
\usepackage{array}
\usepackage{multirow}
\usepackage{gensymb}
\usepackage{tabularx}
\usepackage{extarrows}
\usepackage{booktabs}
\usetikzlibrary{fadings}
\usetikzlibrary{patterns}
\usetikzlibrary{shadows.blur}
\usetikzlibrary{shapes}

\usepackage[T1]{fontenc}

\usepackage[utf8]{inputenc}

\usepackage{microtype}

\usepackage{inconsolata}

\title{Robust Domain Adaptation for Pre-trained Multilingual Neural Machine Translation Models} 

    \author{Mathieu Grosso, {\bf Pirashanth Ratnamogan,} {\bf Alexis Mathey,} \\ {\bf William Vanhuffel,} {\bf Michael Fotso Fotso}\\
BNP Paribas \\
\texttt{(mathieu.grosso, pirashanth.ratnamogan, alexis.mathey)@bnpparibas.com} \\ \texttt{(william.vanhuffel,michael.fotsofotso)@bnpparibas.com} }
\begin{document}
\maketitle

\begin{abstract}

Recent literature has demonstrated the potential of multilingual Neural Machine Translation (mNMT) models. However, the most efficient models are not well suited to specialized industries. In these cases, internal data is scarce and expensive to find in all language pairs. Therefore, fine-tuning a mNMT model on a specialized domain is hard. In this context, we decided to focus on a new task: \textit{Domain Adaptation of a pre-trained mNMT model on a single pair of language} while trying to maintain model quality on generic domain data for all language pairs. The risk of loss on generic domain and on other pairs is high. This task is key for mNMT model adoption in the industry and is at the border of many others. We propose a fine-tuning procedure for the generic mNMT that combines embeddings freezing and adversarial loss. Our experiments demonstrated that the procedure improves performances on specialized data with a minimal loss in initial performances on generic domain for all languages pairs, compared to a naive standard approach (+10.0 BLEU score on specialized data,  -0.01 to -0.5 BLEU on WMT and Tatoeba datasets on the other pairs with M2M100).

\end{abstract}

\section{Introduction}

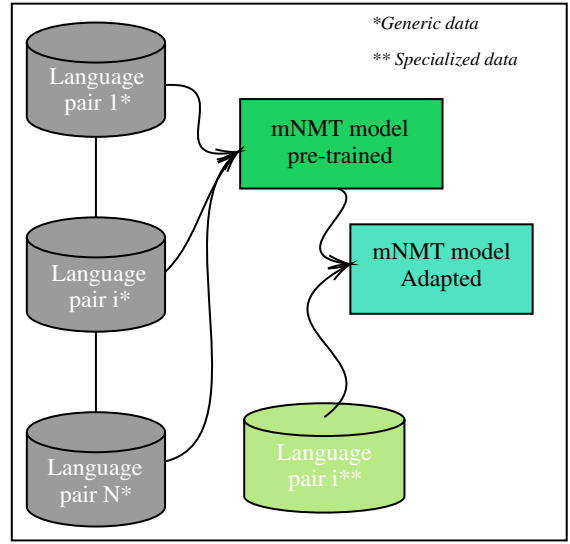
\begin{figure}[h]

\tikzset{every picture/.style={line width=0.75pt}} 

\begin{tikzpicture}[x=0.75pt,y=0.75pt,yscale=-1,xscale=1]

\draw    (44,180) -- (44,199) -- (44,220.5) ;
\draw  [fill={rgb, 255:red, 29; green, 208; blue, 96 }  ,fill opacity=1 ] (116,63) -- (217,63) -- (217,108) -- (116,108) -- cycle ;
\draw    (78,56) .. controls (117.4,53.04) and (73.36,96.18) .. (113.12,89.82) ;
\draw [shift={(115,89.5)}, rotate = 169.46] [color={rgb, 255:red, 0; green, 0; blue, 0 }  ][line width=0.75]    (10.93,-3.29) .. controls (6.95,-1.4) and (3.31,-0.3) .. (0,0) .. controls (3.31,0.3) and (6.95,1.4) .. (10.93,3.29)   ;
\draw    (73,154.5) .. controls (94.56,139.8) and (95.95,109.73) .. (113.88,90.65) ;
\draw [shift={(115,89.5)}, rotate = 135] [color={rgb, 255:red, 0; green, 0; blue, 0 }  ][line width=0.75]    (10.93,-3.29) .. controls (6.95,-1.4) and (3.31,-0.3) .. (0,0) .. controls (3.31,0.3) and (6.95,1.4) .. (10.93,3.29)   ;
\draw    (74,246) .. controls (131.42,240.06) and (76.13,121.9) .. (113.83,90.43) ;
\draw [shift={(115,89.5)}, rotate = 143.13] [color={rgb, 255:red, 0; green, 0; blue, 0 }  ][line width=0.75]    (10.93,-3.29) .. controls (6.95,-1.4) and (3.31,-0.3) .. (0,0) .. controls (3.31,0.3) and (6.95,1.4) .. (10.93,3.29)   ;
\draw  [fill={rgb, 255:red, 184; green, 233; blue, 134 }  ,fill opacity=1 ] (198,224.25) -- (198,262.75) .. controls (198,267.31) and (180.09,271) .. (158,271) .. controls (135.91,271) and (118,267.31) .. (118,262.75) -- (118,224.25) .. controls (118,219.69) and (135.91,216) .. (158,216) .. controls (180.09,216) and (198,219.69) .. (198,224.25) .. controls (198,228.81) and (180.09,232.5) .. (158,232.5) .. controls (135.91,232.5) and (118,228.81) .. (118,224.25) ;
\draw  [fill={rgb, 255:red, 80; green, 227; blue, 194 }  ,fill opacity=1 ] (171,126) -- (262,126) -- (262,171) -- (171,171) -- cycle ;
\draw   (2,15) -- (279,15) -- (279,285) -- (2,285) -- cycle ;
\draw  [fill={rgb, 255:red, 155; green, 155; blue, 155 }  ,fill opacity=1 ] (78.5,230.56) -- (78.5,267.94) .. controls (78.5,273.49) and (63.05,278) .. (44,278) .. controls (24.95,278) and (9.5,273.49) .. (9.5,267.94) -- (9.5,230.56)(78.5,230.56) .. controls (78.5,236.12) and (63.05,240.63) .. (44,240.63) .. controls (24.95,240.63) and (9.5,236.12) .. (9.5,230.56) .. controls (9.5,225.01) and (24.95,220.5) .. (44,220.5) .. controls (63.05,220.5) and (78.5,225.01) .. (78.5,230.56) -- cycle ;
\draw    (44,82) -- (44,122.5) ;
\draw    (165,108) .. controls (182.73,114.9) and (133.51,141.19) .. (169.31,145.8) ;
\draw [shift={(171,146)}, rotate = 185.86] [color={rgb, 255:red, 0; green, 0; blue, 0 }  ][line width=0.75]    (10.93,-3.29) .. controls (6.95,-1.4) and (3.31,-0.3) .. (0,0) .. controls (3.31,0.3) and (6.95,1.4) .. (10.93,3.29)   ;
\draw  [fill={rgb, 255:red, 155; green, 155; blue, 155 }  ,fill opacity=1 ] (78.5,132.56) -- (78.5,169.94) .. controls (78.5,175.49) and (63.05,180) .. (44,180) .. controls (24.95,180) and (9.5,175.49) .. (9.5,169.94) -- (9.5,132.56)(78.5,132.56) .. controls (78.5,138.12) and (63.05,142.63) .. (44,142.63) .. controls (24.95,142.63) and (9.5,138.12) .. (9.5,132.56) .. controls (9.5,127.01) and (24.95,122.5) .. (44,122.5) .. controls (63.05,122.5) and (78.5,127.01) .. (78.5,132.56) -- cycle ;
\draw  [fill={rgb, 255:red, 155; green, 155; blue, 155 }  ,fill opacity=1 ] (78.5,34.56) -- (78.5,71.94) .. controls (78.5,77.49) and (63.05,82) .. (44,82) .. controls (24.95,82) and (9.5,77.49) .. (9.5,71.94) -- (9.5,34.56)(78.5,34.56) .. controls (78.5,40.12) and (63.05,44.63) .. (44,44.63) .. controls (24.95,44.63) and (9.5,40.12) .. (9.5,34.56) .. controls (9.5,29.01) and (24.95,24.5) .. (44,24.5) .. controls (63.05,24.5) and (78.5,29.01) .. (78.5,34.56) -- cycle ;
\draw    (158,223) .. controls (197.6,193.3) and (113.71,177.32) .. (169.27,146.93) ;
\draw [shift={(171,146)}, rotate = 152.28] [color={rgb, 255:red, 0; green, 0; blue, 0 }  ][line width=0.75]    (10.93,-3.29) .. controls (6.95,-1.4) and (3.31,-0.3) .. (0,0) .. controls (3.31,0.3) and (6.95,1.4) .. (10.93,3.29)   ;

\draw (18,145) node [anchor=north west][inner sep=0.75pt]  [font=\footnotesize] [align=left] {\begin{minipage}[lt]{39.03pt}\setlength\topsep{0pt}
\begin{center}
\textcolor[rgb]{1,1,1}{Language}\\\textcolor[rgb]{1,1,1}{ pair i*}
\end{center}

\end{minipage}};
\draw (130,72) node [anchor=north west][inner sep=0.75pt]  [font=\footnotesize] [align=left] {\begin{minipage}[lt]{51.54pt}\setlength\topsep{0pt}
\begin{center}
mNMT model\\pre-trained
\end{center}

\end{minipage}};
\draw (129,235) node [anchor=north west][inner sep=0.75pt]  [font=\footnotesize,color={rgb, 255:red, 255; green, 255; blue, 255 }  ,opacity=1 ] [align=left] {\begin{minipage}[lt]{41.29pt}\setlength\topsep{0pt}
\begin{center}
Language \\pair i**
\end{center}

\end{minipage}};
\draw (175,134) node [anchor=north west][inner sep=0.75pt]  [font=\small] [align=left] {\begin{minipage}[lt]{60.19pt}\setlength\topsep{0pt}
\begin{center}
mNMT model \\Adapted
\end{center}

\end{minipage}};
\draw (17,46) node [anchor=north west][inner sep=0.75pt]  [font=\footnotesize] [align=left] {\begin{minipage}[lt]{39.03pt}\setlength\topsep{0pt}
\begin{center}
\textcolor[rgb]{1,1,1}{Language}\\\textcolor[rgb]{1,1,1}{ pair 1*}
\end{center}

\end{minipage}};
\draw (16,243) node [anchor=north west][inner sep=0.75pt]  [font=\footnotesize] [align=left] {\begin{minipage}[lt]{39.03pt}\setlength\topsep{0pt}
\begin{center}
\textcolor[rgb]{1,1,1}{Language}\\\textcolor[rgb]{1,1,1}{ pair N*}
\end{center}

\end{minipage}};

\draw (180,20) node [anchor=north west][inner sep=0.75pt]   [align=left] {\textit{{\scriptsize *Generic data }}\\{\scriptsize \textit{** Specialized data }}};

\end{tikzpicture}

\caption{Domain Adaptation of a Pre-trained mNMT} 
\label{fig:Domain_adaptation}
\end{figure}

Building a NMT model supporting multiple language pairs is an active and emerging area of research \citep{NLLBteam:22,Fan:20,Tang:20}. Multilingual NMT(mNMT) uses a single model that supports translation in multiple language pairs. Multilingual models have several advantages over their bilingual counterparts \citep{Azivazhagan:19}. This modeling proves to be both efficient and effective as it reduces the operational cost (a single model is deployed for all language pairs) and improves translation performances, especially for low-resource languages.

All these advantages make mNMT models interesting for real-world applications. However, they are not suitable for specialized industries that require domain-specific translation. Training a model from scratch or fine-tuning all the pairs of a pre-trained mNMT model is almost impossible for most companies as it requires access to a large number of resources and specialized data. That said, fine-tuning a single pair of a pre-trained mNMT model in a specialized domain seems possible. Ideally this domain adaptation could be learned while sharing parameters from old ones, without suffering from catastrophic forgetting \citep{mccloskey:89}. This is rarely the case. The risk of degrading performance on old pairs is high due to the limited available data from the target domain and to the extremely high complexity of the pre-trained model. \textbf{In our case, overfitting on fine-tuning data means that the model might not even be multilingual anymore}

In this context, this article focuses on a new real-world oriented task \textbf{fine-tuning a pre-trained mNMT model in a single pair of language on a specific domain without losing initial performances on the other pairs and generic data}. 
Our research focuses on fine-tuning two state-of-the-art pre-trained multilingual mNMT freely available: M2M100 \citep{Fan:20} and mBART50 \citep{Tang:20} which both provide high performing BLEU scores and translate up to 100 languages.


 \textbf{We explored multiple approaches for this domain adaptation.}  Our experiments were made on English to French data from medical domain\footnote{https://opus.nlpl.eu/EMEA-v3.php}. This paper shows that fine-tuning a pre-trained model with initial layers freezing, for a few steps and with a small learning rate is the best performing approach.

It is organized as follows : firstly, we introduce standard components of modern NMT, secondly we describe related works, thirdly we present our methods. We finally systematically study the impact of some state-of-the-art fine-tuning methods  and present our results.

Our main contributions can be separated into 2 parts: 
\begin{itemize}
    \item Defining a new real-world oriented task that focuses on  domain adaptation and catastrophic forgetting on multilingual NMT models
    \item Defining a procedure that allows to finetune a pre-trained generic model on a specific domain
\end{itemize}

\section{Background}

\subsection{Neural Machine Translation}

Neural Machine Translation (NMT) has become the dominant field of machine translation. It studies how to automatically translate from one language to another using neural networks. 

Most NMT systems are trained using Seq2Seq architectures \citep{sutskever2014sequence,cho2014learning} by maximizing the prediction of the target sequence $V_T = (v_{1}, \ldots, v_{T})$, given the source sentence $W_S= (w_{1}, \ldots, w_{S})$: \\
\\
$$P\left(v_{1}, \ldots, v_{T} \mid w_{1}, \ldots, w_{S}\right)$$
\\

Today the best performing Seq2Seq architecture for NMT is based on Transformers \citep{Vaswani:17} architecture. They are built on different layers among which the multi-head attention and the feed-forward layer. These are applied sequentially and are both followed by a residual connection \citep{He:15} and layer normalization \citep{Ba:16}. \\ Although powerful, traditional NMT only translates from one language to another with a high computational cost compared to its statistical predecessor. It has been shown that a simple language token can condition the network to translate a sentence in any target language from any source language \citep{johnson-etal-2017-googles}. It allows to create multilingual models that can translate between multiple languages. Using previous notation the multilingual model adds the condition on target language in the previous modeling

$$P\left(v_{1}, \ldots, v_{T} \mid w_{1}, \ldots, w_{S}, \ell\right) $$
\\
\text{where } $\ell$ \text{is the target language}.

\subsection{Transfer Learning}

Transfer learning is a key topic in Natural Language Processing \citep{Devlin:18,liu:19}. 
It is based on the assumption that pre-training a model on a large set of data in various tasks will help initialize a network trained on another task where data is scarce.

It is already a key area of research in NMT where large set of generic data are freely available (news, common crawl, ...). However, real-world applications require specialized models. In-domain data is rare and more costly to gather for industries (finance, legal, medical, ...) making specialized models harder to train. It is even more true for multilingual model. 

In our work, we study how we can adapt a mNMT model on a specific domain by fine-tuning on only one language pair, without losing too much generality for all language pairs.

\section {Related works}

\subsection{Multilingual Neural Machine Translation}

While initial research on NMT started with bilingual translation systems \citep{sutskever2014sequence, cho2014learning,luong2015effective,Yang:20}, it has been shown that the NMT framework is extendable to multilingual models \citep{dong-etal-2015-multi,Firat:16,johnson-etal-2017-googles, dabre2020survey} mNMT has seen a sharp increase in the number of publications, since it is easily extendable and it allows both end-to-end modeling and cross lingual language representation \citep{conneau2017word,linger2020batch, conneau2019unsupervised}. 

Competitive multilingual models have been released and open sourced. mBART \citep{liu:19} first, was trained following the BART \citep{Lewis:19} objective before being finetuned on an English-centric multilingual dataset \citep{Tang:20}. 
M2M100 \citep{Fan:20} scaled large transformer layers \citep{Vaswani:17} with a lot of mined data in order to create a mNMT without using English as pivot, that can perform translation between any pairs among 100 languages. 
More recently, NLLB was released \citep{NLLBteam:22}, extending the M2M100 framework to 200 languages. Those models are extremely competitive as they have similar performance to their bilingual counterpart while allowing a pooling of training and resources.

Our experiments will rely on M2M100 and mBART but it can be generalized to any new pre-trained multilingual model \citep{NLLBteam:22}.

\subsection{Domain Adaptation}

Domain Adaptation in the field of NMT is a key real-world oriented task. It aims at maximizing model performances on a certain in-domain data distribution.  Dominant approaches are based on fine-tuning a generic model using either in-domain data only or a mixture of out-of-domain and in-domain data to reduce overfitting \citep{servan2016domain,van2017dynamic}. Many works have extended domain adaptation to multi-domain, where model is finetuned on multiple and different domains \citep{sajjad2017neural,zeng2018multi,mghabbar2020building}. 
\\ However, to the best of our knowledge, our work is the first exploring domain adaptation in the context of recent pre-trained multilingual neural machine translation systems, while focusing on keeping the model performant in out-of-domain data in all languages.

\subsection{Learning without forgetting}

Training on a new task or new data without losing past performances is a generic machine learning task, named Learning without forgetting \citep{Li:16}. 

Limiting pre-trained weights updates using either trust regions or adversarial loss is a recent idea that has been used to improve training stability in both natural language processing and computer vision \citep{zhu2019freelb,Jiang:20,aghajanyan2020better}. These methods haven't been explored in the context of NMT but are key assets that demonstrated their capabilities on other NLP tasks (Natural Language Inference in particular). 
Our work will apply a combination of those methods to our task.

\subsection{Zero Shot Translation}

MNMT has shown the capability of direct translation between language pairs unseen in training: a mNMT system can automatically translate between unseen pairs without any direct supervision, as long as both source and target languages were included in the training data \cite{johnson-etal-2017-googles}. However, prior works \citep{johnson-etal-2017-googles, Firat:16, Arivazhagan:19} showed that the quality of zero-shot NMT significantly lags behind pivot-based translation \citep{Gu:19}.
Based on these ideas, some paper \citep{liu:21} have focused on training a mNMT model supporting the addition of new languages by relaxing the correspondence between input tokens and encoder representations, therefore improving its zero-shot capacity.
We were interested in using this method as learning less specific input tokens during the finetuning procedure could help our model not to overfit the training pairs. Indeed, generalizing to a new domain can be seen as a task that includes generalizing to an unseen language.

\section{Methods}

Our new real-world oriented task being at the cross-board of many existing task, we applied ideas from current literature and tried to combine different approaches to achieve the best results.

\subsection{Hyperparameters search heuristics for efficient fine-tuning}

We seek to adapt generic multilingual model to a specific task or domain. \citep{cettolo:hal-01157893,Servan:16}. 
Recent works in NMT \citep{Domingo:19} have proposed methods to adapt incrementally a model to a specific domain. 
We continue the training of the generic model on specific data, through several iterations (see Algorithm \ref{alg:params_tuning}). This post-training fine-tuning procedure is done without dropping the previous learning states of the multilingual model. The resulting model is considered as adapted or specialized to a specific domain. 
We want to avoid the model to suffer from forgetting on generic domain and pairs. To this end, we include different methods in this fine-tuning, that have been mentioned in the literature. These methods includes in particular choosing a small learning rate \citep{Howard&Ruder:18}, a triangular learning schedule \citep{Houlsby:19}, reducing the number of steps and freezing some of the layers\citep{Stickland:19}.

\subsection{Smoothness-inducing Adversarial Regularizer}

We seek to reduce the loss on generic domain and other pairs.  Indeed, due to limited data resources from downstream tasks and the extremely large capacity of pre-trained models, aggressive fine-tuning often causes the adapted model to overfit the data of downstream tasks and forget the knowledge of the pre-trained model. To this end, we added a Smoothness-inducing Adversarial Regularization (SMART) term during the fine-tuning \citep{Jiang:20}. Models fine-tuned on GLUE task with SMART approach outperform even the strongest pre-trained baseline on all 8 tasks. Comparing with BERT \citep{Devlin:18} and RoBERTa \citep{liu:19}, BERT$_{SMART}$ and RoBERTa$_{SMART}$ are performing better by a big margin. This approach gives a smoothness-inducing property to the model $f$. This is helpful to prevent overfitting and to improve generalization on low resource target domain for a certain task. Therefore, adding it to our task should avoid overfitting on the new domain.

Given the model $f(.;\theta)$ and $n$ data points of the target task denoted by $\left\{\left(x_{i}, y_{i}\right)\right\}_{i=1}^{n}$, where $x_{i}$'s denote the embeddings of the input sentences, given by the first embedding layer of the language model and $y_{i}$'s are the associated labels, SMART is adding a regularization term $\mathcal{R}_{\mathrm{s}}(\theta)$ to the canonical optimisation loss below:

\begin{equation}
\label{eq:1}
   \min_\theta(\mathcal{F}(\theta)) = \mathcal{L}(\theta) + \lambda_{\mathrm{s}}\mathcal{R}_{\mathrm{s}}(\theta)
\end{equation}
where $\mathcal{L}(\theta)$ is the loss function defined as 
\begin{equation}
    \mathcal{L}(\theta)=\frac{1}{n} \sum_{i=1}^{n} \ell\left(f\left(x_{i} ; \theta\right), y_{i}\right)
\end{equation}
and $\ell(\cdot, \cdot)$ is the loss function depending on the target task, $\lambda_{\mathrm{s}}>0$ is a tuning parameter, and $\mathcal{R}_{\mathrm{s}}(\theta)$ is the smoothness-inducing adversarial regularizer. Here we define $\mathcal{R}_{\mathrm{s}}(\theta)$ as
\begin{equation}
    \mathcal{R}_{\mathrm{s}}(\theta)=\frac{1}{n} \sum_{i=1}^{n} \max _{\left\|\bar{x}_{i}-x_{i}\right\|_{p} \leq \epsilon} \ell_{\mathrm{s}}\left(f\left(\bar{x}_{i} ; \theta\right), f\left(x_{i} ; \theta\right)\right)
\end{equation}

where $\epsilon>0$ is a tuning parameter. Since NMT is a classification tasks, $f(; \theta)$ outputs a probability simplex and $\ell_{\mathrm{s}}$ is chosen as the symmetrized KL-divergence, i.e.,
$$
\ell_{\mathrm{s}}(P, Q)=\mathcal{D}_{\mathrm{KL}}(P \| Q)+\mathcal{D}_{\mathrm{KL}}(Q \| P)
$$

\subsection{Enabling the model to learn less aggressive input tokens}


We seek at reducing the loss of performances on the pairs learned during the pre-training of the model. A factor causing a too important language-specific representation is the positional correspondence to input tokens \citep{liu:21}. Relaxing it should help the model learn the new domain while not focusing too much on the language representation. Recent advances in mNMT showed that we can reduce the positional correspondence learned from the input tokens seen during training thanks to Positional Disentangling Encoder (PDE) \citep{liu:21}. PDE  corresponds to removing some of the residual connections of the model architecture. PDE is reported to beat by +18.5 BLEU models that do not use it on zero shot translation pairs while retaining quality on supervised directions \citep{liu:21}. 
Doing this during the domain adaptation fine-tuning helps to learn less specific input tokens (since we train only from English to French). Therefore, adapting this method to our domain adaptation training is straightforward and could bring gain in BLEU on language pairs seen during pre-training while not sacrificing performances on the new specific domain.

\section {Experimental Settings}

\subsection{Pre-trained Generic Models used}

We have worked with two pre-trained mNMT models: M2M100 and mBART50 large.

\textbf{M2M100}
is a multilingual encoder-decoder model, based on large Transformer architecture  that can handle 100 languages.  It was trained on a non-English-centric dataset of 7.5B sentences from generic domain, as such it is the first true many-to-many NMT model. To ease the fine-tuning process and due to hardware limitations, we worked with the lightest version released (418M parameters).

\textbf{mBART50}
is a multilingual encoder-decoder model, based on training on an English-centric dataset and on large Transformer architecture that can handle 50 languages. It was trained following the BART objective \citep{Lewis:19}. More formally, the model aims to reconstruct a text that has been previously noised.

We will compare the domain adaptation performance between mBART50 which was trained on English-centric data and M2M100 which was trained on non English-centric data. \\

\subsection{Datasets and preprocessing}
In order to assess the effectiveness of our different domain adaptation strategies, we focused on the medical domain on the \textbf{English to French} using data from the EMEA3\footnote{https://opus.nlpl.eu/EMEA.php} dataset \citep{TIEDEMANN:12}. We used the same preprocessing as the original publications (BPE joint-tokenization from sentencepiece). We split the dataset into a train and a test dataset. We chose to use the first 5.000 sentences for the testing set and 350.000 sentences for the training set.
For the evaluation data on the generic domain, we used generic data from different sources including WMT\footnote{https://opus.nlpl.eu/WMT-News.php} and Tatoeba\footnote{https://github.com/Helsinki-NLP/Tatoeba-Challenge}.
For the evaluation data on the medical domain, we also used EMEA3 dataset in different languages.

\subsection{Detailed Procedure}

We first define a hyperparameters search heuristics procedure. We chose a range of learning rate and trained the model with these values. We set prior threshold between the loss we accept on generic data and the increase we target on medical data. Then apply the procedure in algorithm \ref{alg:params_tuning}. Having done this, we kept best settings (best learning rate and number of steps for given threshold), and tried freezing first layers to reduce the loss on generic domain. We define ${\epsilon}_3$, a threshold between loss on medical domain and gain on generic domain. We reproduce the same procedure and reports our best results. This allows us to find the optimal model $\theta_{opt}$, representing the best compromise between not losing performances on generic data and good adaptation to the medical domain.

\begin{algorithm}[H]
\caption{Hyperparameters search heuristic for domain adaptation using simple fine-tuning Algorithm}\label{alg:params_tuning}
\begin{algorithmic}[1]
\Require{$T$ : the maximum number of steps; $L$ : the number of layers we have frozen; $L_r$: the learning rate, $\epsilon_1$: the threshold for  ${\Delta}_1$ : the difference of BLEU between baseline and adapted model on EN-FR generic domain data, $\epsilon_2$ threshold for $\Delta_2$ : the mean difference of BLEU between baseline and adapted model on all other generic data, $\theta_0$ is the parameters of the pretrained model, $\theta_{opt}$: is the parameters of the model that has optimal value of BLEU on domain and generic.}
\State $ T \gets 100K$
\State $L \gets 1$
\For{$L_r=3e-5,1e-5,...,1e-8$}
\State $\theta_s \gets \theta_0$
\For{$s \gets 1$ to $T$} 
\State $\theta_{s+1} \leftarrow$ AdamUpdate $_{B}\left({\theta}_{s}\right)$
\State Every 2k steps, evaluate model on validation set and compute $\Delta_1$ and $\Delta_2$
\If {$\Delta_1 \leq \epsilon_1 \cup \Delta_2 \leq \epsilon_2$ is true}
\State $\theta_{opt} \gets \theta_{s}$
\Else
\State $\theta_{opt} \gets \theta_{s}$
\State end For loop
\EndIf
\EndFor
\EndFor
\Ensure{$\theta_{opt}$}
\end{algorithmic}
\end{algorithm}

\textbf{M2M100}
We trained M2M100 on the medical EN-FR dataset. We used the adam optimizer ($\beta_{1}=0.9, \beta_{2}=0.98$), label smoothing, a dropout of 0.1 and a weight decay of 0. We applied our hyperparameters search heuristic procedure \ref{alg:params_tuning} to find the best model. We set $\epsilon_1=2, \epsilon_2=1$. On this configuration, optimal results were reported with  a learning rate of 1e-07, freezing the embeddings at the encoder level, and 60K steps.\\
\textbf{mBART50}
We trained mBART50 large on the medical EN-FR dataset. We used the adam optimizer ($\beta_{1}=0.9, \beta_{2}=0.98$), label smoothing, a dropout of 0.3 and a weight decay of 0. Again, we applied our hyperparameters search heuristic procedure to find the best model \ref{alg:params_tuning}. We had to increase the value of $\epsilon_1, \epsilon_2$ since mBART50 tends to forget the generic domain quicker than M2M100. We set $\epsilon_1=4, \epsilon_2=3$. On this configuration, optimal results were reported with a learning rate of $6e-07$, freezing the embeddings at the encoder level and 10K steps.\\
\\
\textbf{SMART:} 
We finetuned the model with the SMART procedure and continue hyperparameters search as in algorithm \ref{alg:params_tuning}.
In Algorithm \ref{alg:smart}, we note $\mathcal{R}_{\mathrm{s}}(\theta)=\frac{1}{\lvert B \rvert} \sum_{x_{i} \in B} \max_{\left\|\bar{x}_{i}-x_{i}\right\| p \leq \epsilon} \ell_{\mathrm{s}}\left(f\left(\bar{x}_{i} ; \theta\right), f\left(x_{i} ; \theta\right)\right)$ and $AdamUpdate$ the ADAM update rule for optimizing equation \ref{eq:1} using the mini-batch $B$. Lastly, we set $T_{\tilde{x}}=1$. For the perturbation, we set $\epsilon=10^{-5}$ and $\sigma=10^{-5}$. The learning rate $\eta$ is set to $10^{-3}$.

\begin{algorithm}[h] 
\caption{Adding SMART to procedure} \label{alg:smart}
\renewcommand{\algorithmicrequire}{\textbf{Input:}}
\renewcommand{\algorithmicensure}{\textbf{Output:}}
\begin{algorithmic}[1]
\Require{$T$ : the total number of iterations; $\mathcal{X}$: the dataset; $\theta_{0}$: the parameter of the pre-trained model; $\sigma^{2}$: the variance of the random initialization for $\bar{x}_{i}$ 's; $T_{\tilde{x}}$: the number of iterations for updating $\bar{x}_{i}$ 's; $\eta$: the learning rate for updating $\bar{x}_{i}$ 's; $\beta$: clipping value.}
\State $\theta_{1} \gets \theta_{0}$
\For{$t \gets 1$ to $T$} 
\State $\bar{\theta}_{s} \gets \theta_{t}$
\State Sample a mini-batch $B$ from $\mathcal{X}$
\State For all $x_{i} \in B$, initialize $\bar{x}_{i} \gets x_{i}+v_{i}$ with $v_{i} \sim \mathcal{N}\left(0,\sigma^{2}I\right)$
\For{$m=1,...,T_{\tilde{x}}$}
\State $\bar{x}_{i} \gets \bar{x}_{i} + \eta\mathcal{R}_{\mathrm{s}}(\bar{\theta}_{s})$
\EndFor
\State $\bar{\theta}_{s+1} \gets AdamUpdate_{B}\left(\bar{\theta}_{s}\right)$
\State $\theta_{t+1} \gets CLIP(\bar{\theta}_{s+1},1-\beta,1+\beta)$
\EndFor
\Ensure{$\theta_T$}
\end{algorithmic}
\end{algorithm}

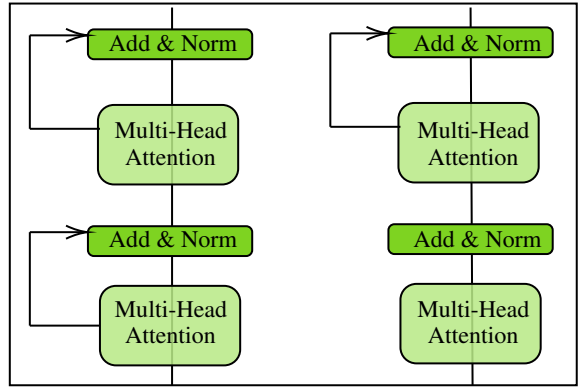
\begin{figure}[H]

\tikzset{every picture/.style={line width=0.75pt}} 

\begin{tikzpicture}[x=0.75pt,y=0.75pt,yscale=-1,xscale=1]

\draw   (20,88) -- (303,88) -- (303,280) -- (20,280) -- cycle ;

\draw    (101,90) -- (101,280) ;
\draw  [fill={rgb, 255:red, 184; green, 233; blue, 134 }  ,fill opacity=0.86 ] (65,238) .. controls (65,233.58) and (68.58,230) .. (73,230) -- (127,230) .. controls (131.42,230) and (135,233.58) .. (135,238) -- (135,262) .. controls (135,266.42) and (131.42,270) .. (127,270) -- (73,270) .. controls (68.58,270) and (65,266.42) .. (65,262) -- cycle ;
\draw  [fill={rgb, 255:red, 184; green, 233; blue, 134 }  ,fill opacity=0.86 ] (64,147) .. controls (64,142.58) and (67.58,139) .. (72,139) -- (126,139) .. controls (130.42,139) and (134,142.58) .. (134,147) -- (134,171) .. controls (134,175.42) and (130.42,179) .. (126,179) -- (72,179) .. controls (67.58,179) and (64,175.42) .. (64,171) -- cycle ;
\draw  [fill={rgb, 255:red, 126; green, 211; blue, 33 }  ,fill opacity=1 ] (59,203) .. controls (59,201.34) and (60.34,200) .. (62,200) -- (138,200) .. controls (139.66,200) and (141,201.34) .. (141,203) -- (141,212) .. controls (141,213.66) and (139.66,215) .. (138,215) -- (62,215) .. controls (60.34,215) and (59,213.66) .. (59,212) -- cycle ;
\draw  [fill={rgb, 255:red, 126; green, 211; blue, 33 }  ,fill opacity=1 ] (59,104) .. controls (59,102.34) and (60.34,101) .. (62,101) -- (138,101) .. controls (139.66,101) and (141,102.34) .. (141,104) -- (141,113) .. controls (141,114.66) and (139.66,116) .. (138,116) -- (62,116) .. controls (60.34,116) and (59,114.66) .. (59,113) -- cycle ;
\draw    (30,250) -- (65,250) ;
\draw    (30,203) -- (30,250) ;
\draw    (30,203) -- (57,203) ;
\draw [shift={(59,203)}, rotate = 180] [color={rgb, 255:red, 0; green, 0; blue, 0 }  ][line width=0.75]    (10.93,-3.29) .. controls (6.95,-1.4) and (3.31,-0.3) .. (0,0) .. controls (3.31,0.3) and (6.95,1.4) .. (10.93,3.29)   ;
\draw    (30,104) -- (57,104) ;
\draw [shift={(59,104)}, rotate = 180] [color={rgb, 255:red, 0; green, 0; blue, 0 }  ][line width=0.75]    (10.93,-3.29) .. controls (6.95,-1.4) and (3.31,-0.3) .. (0,0) .. controls (3.31,0.3) and (6.95,1.4) .. (10.93,3.29)   ;
\draw    (30,104) -- (30,151) ;
\draw    (30,151) -- (65,151) ;
\draw    (250,90) -- (251,280) ;
\draw  [fill={rgb, 255:red, 184; green, 233; blue, 134 }  ,fill opacity=0.86 ] (215,237) .. controls (215,232.58) and (218.58,229) .. (223,229) -- (277,229) .. controls (281.42,229) and (285,232.58) .. (285,237) -- (285,261) .. controls (285,265.42) and (281.42,269) .. (277,269) -- (223,269) .. controls (218.58,269) and (215,265.42) .. (215,261) -- cycle ;
\draw  [fill={rgb, 255:red, 184; green, 233; blue, 134 }  ,fill opacity=0.86 ] (214,146) .. controls (214,141.58) and (217.58,138) .. (222,138) -- (276,138) .. controls (280.42,138) and (284,141.58) .. (284,146) -- (284,170) .. controls (284,174.42) and (280.42,178) .. (276,178) -- (222,178) .. controls (217.58,178) and (214,174.42) .. (214,170) -- cycle ;
\draw  [fill={rgb, 255:red, 126; green, 211; blue, 33 }  ,fill opacity=1 ] (209,202) .. controls (209,200.34) and (210.34,199) .. (212,199) -- (288,199) .. controls (289.66,199) and (291,200.34) .. (291,202) -- (291,211) .. controls (291,212.66) and (289.66,214) .. (288,214) -- (212,214) .. controls (210.34,214) and (209,212.66) .. (209,211) -- cycle ;
\draw  [fill={rgb, 255:red, 126; green, 211; blue, 33 }  ,fill opacity=1 ] (209,103) .. controls (209,101.34) and (210.34,100) .. (212,100) -- (288,100) .. controls (289.66,100) and (291,101.34) .. (291,103) -- (291,112) .. controls (291,113.66) and (289.66,115) .. (288,115) -- (212,115) .. controls (210.34,115) and (209,113.66) .. (209,112) -- cycle ;
\draw    (180,103) -- (207,103) ;
\draw [shift={(209,103)}, rotate = 180] [color={rgb, 255:red, 0; green, 0; blue, 0 }  ][line width=0.75]    (10.93,-3.29) .. controls (6.95,-1.4) and (3.31,-0.3) .. (0,0) .. controls (3.31,0.3) and (6.95,1.4) .. (10.93,3.29)   ;
\draw    (180,103) -- (180,150) ;
\draw    (180,150) -- (215,150) ;

\draw (69,236) node [anchor=north west][inner sep=0.75pt]  [font=\small] [align=left] {\begin{minipage}[lt]{44.97pt}\setlength\topsep{0pt}
\begin{center}
{\small Multi-Head }\\{\small Attention}
\end{center}

\end{minipage}};
\draw (69,146) node [anchor=north west][inner sep=0.75pt]  [font=\small] [align=left] {\begin{minipage}[lt]{44.97pt}\setlength\topsep{0pt}
\begin{center}
{\small Multi-Head }\\{\small Attention}
\end{center}

\end{minipage}};
\draw (68,201) node [anchor=north west][inner sep=0.75pt]  [font=\small] [align=left] {{\footnotesize Add \& Norm}};
\draw (68,102) node [anchor=north west][inner sep=0.75pt]  [font=\small] [align=left] {{\footnotesize Add \& Norm}};
\draw (220,236) node [anchor=north west][inner sep=0.75pt]  [font=\small] [align=left] {\begin{minipage}[lt]{44.97pt}\setlength\topsep{0pt}
\begin{center}
{\small Multi-Head }\\{\small Attention}
\end{center}

\end{minipage}};
\draw (220,146) node [anchor=north west][inner sep=0.75pt]  [font=\small] [align=left] {\begin{minipage}[lt]{44.97pt}\setlength\topsep{0pt}
\begin{center}
{\small Multi-Head }\\{\small Attention}
\end{center}

\end{minipage}};
\draw (220,201) node [anchor=north west][inner sep=0.75pt]  [font=\small] [align=left] {{\footnotesize Add \& Norm}};
\draw (220,102) node [anchor=north west][inner sep=0.75pt]  [font=\small] [align=left] {{\footnotesize Add \& Norm}};

\end{tikzpicture}

\caption{PDE Illustration: Removing Residual Connections on encoder block} 
\label{fig:PDE}
\end{figure}

\textbf{PDE}
Finally, we define PDE. It consists in applying Algorithm \ref{alg:params_tuning} and then removing first all the residual connection in the penultimate Encoder layers \citep{chen:22}, then we try removing only the attention layer residual connections (figure \ref{fig:PDE}).

\section{Results and Analysis}

\subsection{Hyperparameters search heuristic}
\subsubsection{Main Results}

\begin{figure}[h]
\centering
\includegraphics[scale=0.3]{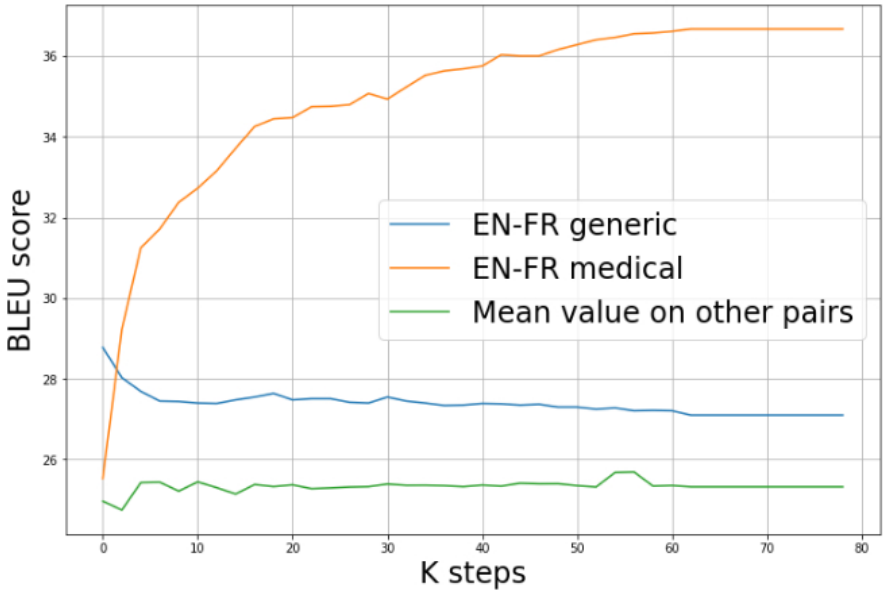}
\caption{Domain Adaptation (Medical Domain EN-FR) of M2M100}
\label{fig:M2M_main}
\end{figure}

\textbf{M2M100}
As shown in table \ref{tab:main_res_table} we reached more than 9.00 increase of BLEU score on the medical dataset without sacrificing performance on generic domain, the loss is not important on most of the pairs (between 0.01 and 0.2). In figure \ref{fig:M2M_main}, we see that the mean results is rather stable and that the BLEU on generic English to French data does not decrease a lot (around -1.5 BLEU). The model converges after 60K steps so we stop training. 

\begin{figure}[h]
\centering
\includegraphics[scale=0.3]{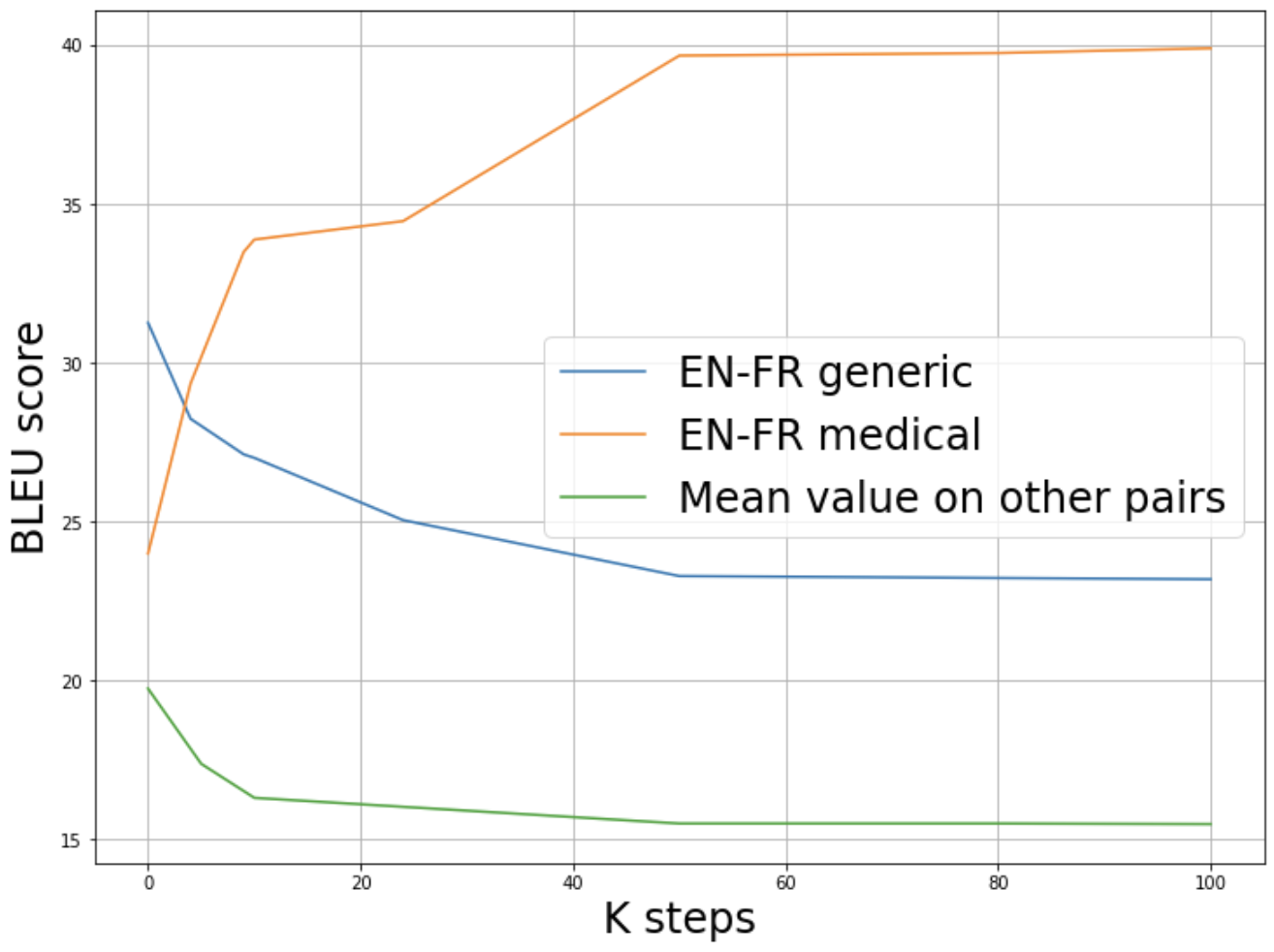}
\caption{Domain Adaptation (Medical Domain EN-FR) of mBART50}
\label{fig:mBART_main}
\end{figure}

\textbf{mBART50}
Again we reach more than 9.00 BLEU increase (Figure \ref{fig:mBART_main}). We observe that after 50K steps mBART50 starts converging around 40.00 BLEU, yet we decided to stop domain adaptation training sooner than with M2M100 as a trade-off between good performance on the EN-FR medical domain and loss of performance on the generic domain. Globally, we achieved better results with M2M100 than mBART50.

\subsubsection{Catastrophic forgetting with a big learning rate}

We tested several learning rate values and we report here our results with a bigger learning rate ($3e-5$). For both models, it led to a catastrophic forgetting on the non-finetuned pairs along with a huge performance increase on the EN-FR Medical dataset, reaching a higher BLEU on the Medical dataset. We decided to focus on a smaller learning rate as a trade-off between loss on generic domain and gain on the medical domain.


\begin{figure}[H]
\centering
\includegraphics[scale=0.3]{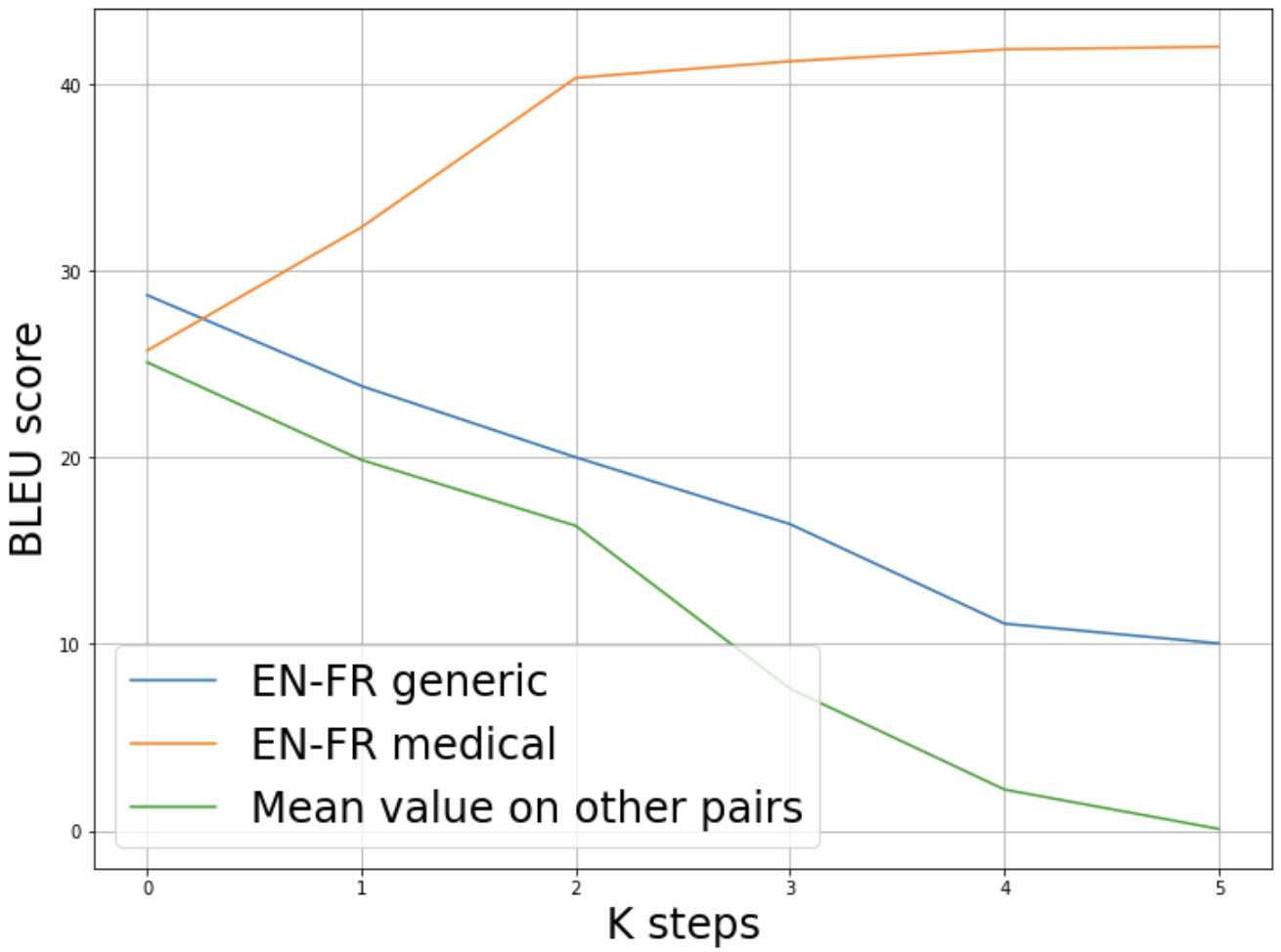}
\caption{Domain Adaptation of M2M100 with big Learning Rate}
\label{fig:cata_M2M100}
\end{figure}

\begin{table*}[h]
\centering
\caption{Global results on domain adaptation of M2M100 and mBART50}
\label{tab:main_res_table}
\resizebox{\linewidth}{!}{%
\begin{tabular}{c|cccc|cccc} 
\toprule
                   & \multicolumn{4}{c}{\textbf{M2M100}}                                                        & \multicolumn{4}{c}{\textbf{mBART50}}                                                        \\
                   & \textbf{Baseline} & \textbf{Finetuned} & \textbf{Finetuned SMART} & \textbf{Finetuned PDE} & \textbf{Baseline} & \textbf{Finetuned} & \textbf{Finetuned SMART} & \textbf{Finetuned PDE}  \\ 
\cmidrule{1-9}
EN-FR medical data & 26.94             & \textbf{36.05}     & 35.93                    & 29.71                  & 23.99             & \textbf{33.87}     & 30.3                     & 26.12                   \\
EN-FR (WMT)        & \textbf{28.63}             & 26.90              & 26.41                    & 25.01                  & \textbf{31.25}            & 27.10              & 26.10                    & 17.56                   \\
Mean results       & 24.97             & \textbf{25.38}         & 25.15                    & 21.70                  & \textbf{19.83}             & 16.85              & 15.1                     & 13.63                   \\ 
\cmidrule{1-9}
DE-EN (WMT)        & 22.42             & \textbf{22.53}          & 22.39                    & 21.15                  & \textbf{26.13}            & 21.85              & 20.64                    & 19.44                   \\
EN-DE (WMT)        & 19.48             & \textbf{19.52}           & 19.21                    & 17.96                  & \textbf{22.72}            & 19.84              & 18.61                    & 16.83                   \\
RU-EN (WMT)        & \textbf{26.3}           & 26.27              & 25.4                     & 24.90                  & \textbf{29.72}             & 24.64              & 23.55                    & 22.47                   \\
FR-DE (WMT)        & \textbf{17.82}            & 17.80              & 17.58                    & 14.69               & \textbf{10.92}           & 8.98               & 7.1                      & 4.30                    \\
EN-FI (WMT)        & 12.51             & \textbf{12.72}          & 12.51                    & 11.74                  & \textbf{13.39}            & 11.23              & 10.27                    & 9.74                    \\
FI-EN (WMT)        & \textbf{23.55}            & 23.18              & 23.39                    & 21.40                  & \textbf{22.10}            & 18.90              & 17.75                    & 16.33                   \\
BG-IT (Tatoeba)    & 26.65             & \textbf{27.54}             & 27.01                    & 26.20                  & *                 & *                  & *                        & *                       \\
DA-TR (Tatoeba)    & 20.22             & \textbf{22.27}            & 21.75                    & 20.23                  & *                 & *                  & *                        & *                       \\
PL-RU (Tatoeba)    & \textbf{33.79}             & 33.72              & 33.68                    & 29.69                  & \textbf{14.45}         & 10.49              & 10.34                    & 9.87                    \\
PT-ES (Tatoeba)    & 51.49             & 51.98              & \textbf{52.54 }                   & 50.34                  & \textbf{21.57}             & 18.87              & 16.80                    & 12.54                   \\
JA-ES (Tatoeba)    & 20.55             & \textbf{21.66}              & 21.58                    & 20.30                  & \textbf{17.55}             & 15.42              & 13.27                    & 11.12                   \\
\bottomrule
\end{tabular}
}
\end{table*}

\subsection{SMART}

\begin{figure}[h]
\centering
\includegraphics[scale=0.3]{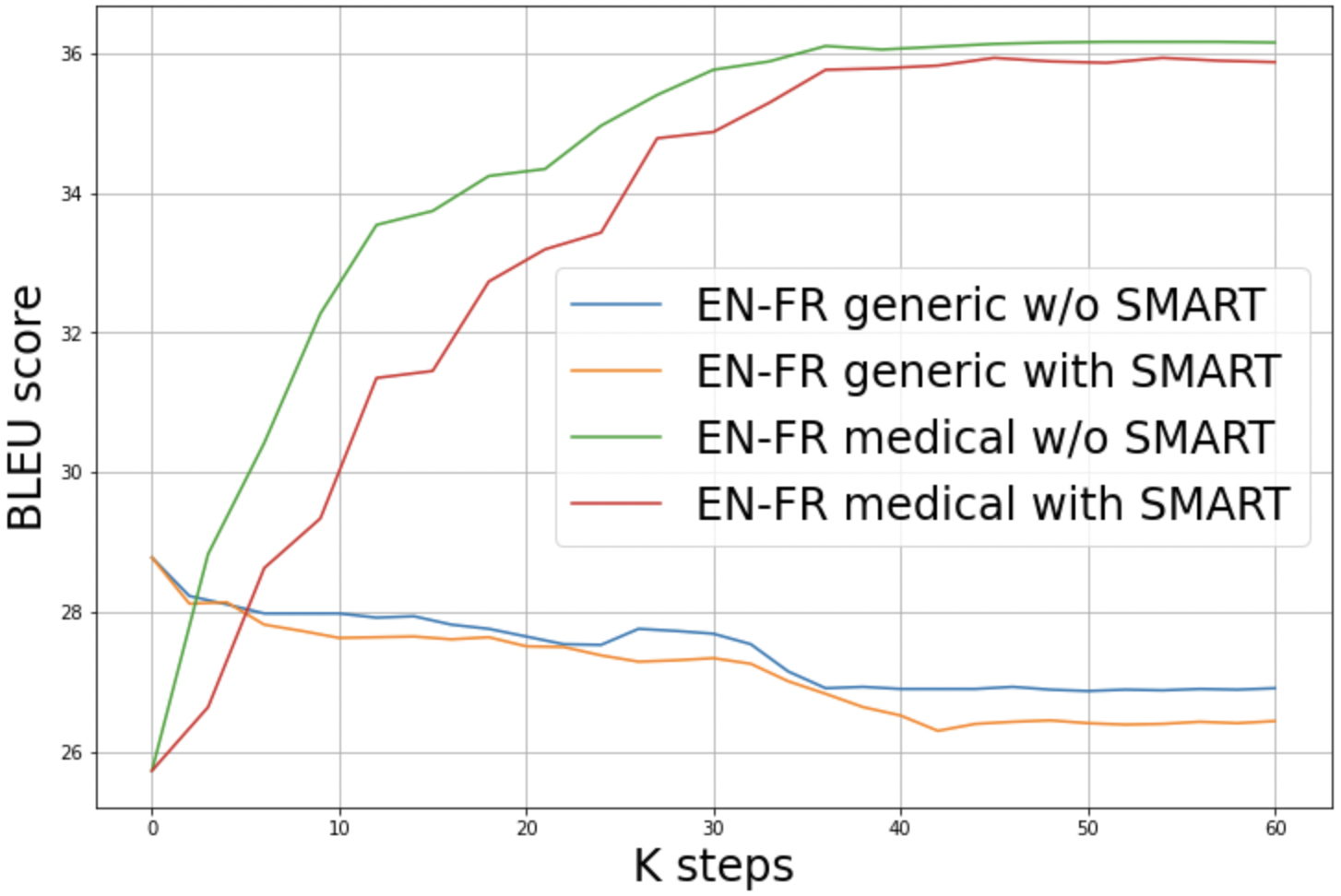}
\caption{Adding SMART to M2M100 Domain Adaptation training}
\label{fig:smart_M2M100}
\end{figure}

We have reported our fine-tuning results for M2M100 and mBART50 with SMART in Table \ref{tab:main_res_table}. 

Our goal with SMART was to reach a higher BLEU score on the generic domain data without sacrificing performances on the medical dataset. 
In Table \ref{tab:main_res_table}, we note a good increase in BLEU score. Moreover, we have noted that the BLEU change less when moving learning rate in a reasonable range compared to the other methods that are extremely sensitive to hyperparameters. In this context, SMART is useful in order to \textbf{achieve quick adaptation of a mNMT model to a new domain. It makes domain adaptation procedure more consistent}. Therefore, SMART training procedure allows efficient and robust domain adaptation. 
However if exploring a large scale of hyper parameters if feasible simple fine-tuning procedure like in Algorithm \ref{alg:params_tuning} can provide better results  as shown in Table \ref{tab:main_res_table}.

\subsection{PDE}
We seek at reducing the loss of performances on the pairs learned during pre-training of the model (and that are not used during the post-training domain adaptation). Relaxing the correspondence to the input tokens learned during Domain Adaptation. Fine-tuning was supposed to help learning less specific input tokens and therefore the model would be less likely to forgot all the pretrained pairs. 
As expected, the model learned less aggressive input tokens and do not overfit on English input tokens. 
However, in practice this does not seem to work well. Indeed, the model is also likelier to forget the pretrained input tokens making this method unfit to our procedure. Using PDE a posteriori (during fine-tuning) seems to be inefficient, since the model is performing worse on all pairs and not only on the English pairs.

We report our results in table \ref{tab:main_res_table}.

\subsection{Analysis}

\subsubsection{Zero-shot Domain Adaptation on other pairs}

We challenged the approach on domain adaptation on languages unseen during the post-training on the medical domain using EMEA3 dataset available on other languages.
Table \ref{tab:domain_ad} shows that for M2M100 all BLEU scores are increasing, moreover the pairs that implies either English or French are particularly benefiting from this domain adaptation. On mBART50, we also note improvements, first the loss is less important than on generic dataset for the pairs that do not include French as output showing that the model is learning a bit. When French is the output, the domain adaptation is working really fine and we see improvements. 
Domain-specific data are often hard to gather, especially for low-ressource pairs. That's why being able to improve the performances on a new domain for several pairs using a domain-specific dataset from a single pair is a very interesting propriety from the mNMT models.

\begin{table}[h]
\resizebox{\columnwidth}{!}{%
\begin{tabular}{l|ll|ll}
\hline
              & \multicolumn{2}{c|}{\textbf{M2M100}} & \multicolumn{2}{c}{\textbf{mBART50}} \\ \hline
              & \textbf{Baseline}   & \textbf{Ours}  & \textbf{Baseline}   & \textbf{Ours}  \\ \hline
EN-FI medical & 12.93               & \textbf{14.83}         & 10.83               & 10.1           \\
DE-PL medical & 12.62               & \textbf{13.6}          & 11.1                & 7.85           \\
FR-IT medical & 23.07               & \textbf{24.62}         & 10.77               & 8.58           \\
EN-ES medical & 32.38               & \textbf{35.15}        & 15.82               & 17.5           \\
ES-IT medical & 25.43               & \textbf{27.06}       & 8.40                & 7.50           \\
ES-FR medical & 24.37               & \textbf{30.64}         & 19.03               & 25.6           \\
LT-PL medical & 12.49               & \textbf{13.8}        & 8.5                 & 7.9            \\
DE-FR medical & 18.85               & \textbf{22.20}         & 13.3                & 18.19          \\
LT-PT medical & 17.44               & \textbf{19.26}          & *                   & *             
\end{tabular}%
}
\caption{Zero shot domain adaptation on medical dataset for other pairs}
\label{tab:domain_ad}
\end{table}

\subsubsection{Comparison of initial pre-trained mNMT models (mBART 50 vs M2M100)}

We investigated why mBART50 was more likely to forget other pairs compared to M2M100.
First, we have worked with the 418M-parameters version of M2M100. This is not the largest M2M100 version released (and certainly not the most optimized) and this could possibly explain the differences. 
Then, another hypothesis is the different dataset used during training of both models. Indeed, mBART50 is trained on English-centric data, and M2M100 is not. Non-English centric models are known to achieve higher BLEU especially on low resource data \citep{Fan:20}. Extending this study to domain adaptation, we believe non-English-centric models might be more robust to domain adaptation. We noted that when fine-tuning mBART50 with a bigger learning rate, the first pairs to be forgotten are the non-English ones. Testing this hypothesis on NLLB might be useful.

\section{Conclusion and Discussion}

In this paper, we propose a study of robust domain adaptation approaches on mNMT models where in-domain data is available only for a single language pair. Best performing approach combines embedding freezing and simple fine-tuning with good hyperparameters. This approach shows good improvements with few in-domain data on all language pairs. The framework effectively avoids overfitting and aggressive forgetting on out-of-domain generic data while quickly adapting to in-domain data. We demonstrate that this could be a solution for incremental adaptation of mNMT models. 
Finally our work is a call for more research in domain adaptation for multilingual models as it is key for real-world applications.

\section{Limitations}

This study was limited by hardware issues. We did not have the possibility to fine-tune on M2M100 large version (12B parameters) that requires 64 GB of VRAM. 

Testing our results with a larger version of M2M100 might be interesting. 

Also,  our study focused
on two pre-trained multilingual neural machine
translation models. However, many others exist
and will be released \citep{NLLBteam:22}.
We think that our work is generic enough to be applied on other pre-trained models but extensive
experiments on these new models should be carried out.

Finally, the work has been realised on English to French data. We showed domain adaptation is possible for languages with English morphology and tested the impact of this training on many different languages morphology (Japanese, English, Russian, ...). Applying domain adaptation training on other morphology languages and on other domains is also an area to investigate.

\section{Ethics Statement}
The dataset was gathered on OPUS and is largely open-sourced. It was released by \cite{TIEDEMANN:12} and we have downloaded it from OPUS website. We have reviewed the dataset and have not noted any issue with these data. They are very specific to health domain and therefore are not inappropriate. The dataset does not deal with demographic or identity characteristics. 

Moreover, these experiments were made using only 2 GPUs and training were relatively short. Given the urgency of addressing climate change, we  believe our domain adaptation procedure could help have high-performing mNMT models at small carbon and energy costs. Moreover, SMART framework allows for quicker research of the right hyperparameters, therefore reducing even further the number of experiments and the carbon costs of our method. 

\section{Acknowledgments}

The authors would also like to thank Mr. Baoyang Song, Mr. Laurent Lam, Mr. Laatiri Seif Eddine from BNP Paribas for their valuable comments and suggestions. The work is supported by BNP Paribas.

\bibliography{anthology,custom}
\bibliographystyle{acl_natbib}

\end{document}